\documentclass[twoside,11pt]{article}

\usepackage{blindtext}
\usepackage{wrapfig}

\usepackage[preprint]{jmlr2e}

\usepackage{lastpage}
\jmlrheading{XX}{2023}{1-\pageref{LastPage}}{XX/XX; Revised XX/XX}{XX/XX}{XX-XXXX}{Systems Neural Engineering Lab}

\ShortHeadings{\texttt{lfads-torch}}{Sedler and Pandarinath}
\firstpageno{1}

\begin{document}

\title{\texttt{lfads-torch}: A modular and extensible implementation of latent factor analysis via dynamical systems}

\author{\name Andrew R. Sedler \email arsedle@emory.edu \\
       \addr Center for Machine Learning and Department of Biomedical Engineering\\
       Georgia Institute of Technology and Emory University -- Atlanta, GA
       \AND
       \name Chethan Pandarinath \email cpandar@emory.edu \\
       \addr Center for Machine Learning and Department of Biomedical Engineering\\
       Georgia Institute of Technology and Emory University -- Atlanta, GA}

\editor{TBD}

\maketitle

\begin{abstract}
Latent factor analysis via dynamical systems (LFADS) is an RNN-based variational sequential autoencoder that achieves state-of-the-art performance in denoising high-dimensional neural activity for downstream applications in science and engineering. Recently introduced variants and extensions continue to demonstrate the applicability of the architecture to a wide variety of problems in neuroscience. Since the development of the original implementation of LFADS, new technologies have emerged that use dynamic computation graphs, minimize boilerplate code, compose model configuration files, and simplify large-scale training. Building on these modern Python libraries, we introduce \texttt{lfads-torch}---a new open-source implementation of LFADS that unifies existing variants and is designed to be easier to understand, configure, and extend. Documentation, source code, and issue tracking are available at: \href{https://github.com/arsedler9/lfads-torch}{\texttt{https://github.com/arsedler9/lfads-torch}}.
\end{abstract}


\section{Introduction}
Modern technologies for monitoring neural systems include a wide range of modalities, many of which record noisy, high-dimensional observations. These data are often processed and denoised to yield representations of activity that are useful for studying neural function, or for decoding or classifying behaviors. Neural population models (NPMs) accomplish this denoising by leveraging spatiotemporal structure in neural recordings, yielding denoised activity patterns on a single-trial basis and millisecond timescale. NPMs based on artificial neural networks offer significant performance and modularity advantages over previous approaches, as architectures and loss functions can easily be modified to support new modeling goals and data modalities. Latent factor analysis via dynamical systems (LFADS) \cite{sussillo2016lfads, pandarinath2018inferring} is one such model that has matured over the the last several years to support automated hyperparameter tuning \cite{keshtkaran2019enabling, keshtkaran2022large}, electromyography (EMG) and calcium imaging modalities \cite{wimalasena2022estimating, zhu2021deep, zhu2022deep}, and stabilization of long-term recordings \cite{karpowicz2022stabilizing}.

The initial implementation of LFADS used TensorFlow 1.5, which relied on a static computation graph and required users to internalize additional concepts like placeholders and sessions. This made modification, debugging, and inspecting intermediate outputs opaque and cumbersome. Additionally, functions like dataset handling (shuffling, batching, etc.), optimization loops, and logging had to be implemented manually. Finally, model configuration was handled using a command-line interface, so alternative architectures had to be implemented using control flow within the code itself. These factors added substantial complexity to the code base, which made further experimentation and development more challenging and less accessible to the larger neuroscience community.

Since the development of the original model, the ecosystem of deep learning technologies has seen significant growth and maturation. PyTorch introduced dynamic computation graphs and fast eager execution, which allow an intuitive model development and debugging workflow \cite{paszke2019pytorch}. PyTorch Lightning virtually eliminated the need for engineering boilerplate and provided a template for principled compartmentalization of modeling code \cite{falcon2019lightning}. Hydra provided a mechanism for composable configurations, allowing models and their components to be defined in separate configuration files and instantiated on the fly \cite{yadan2019hydra}. Finally, Ray's Tune framework greatly simplified large-scale hyperparameter tuning \cite{liaw2018tune}. Together, these technologies provide an opportunity to substantially lower the barriers to modeling neural data with LFADS and further enhance its capabilities.

In this work we introduce \texttt{lfads-torch}, an implementation of LFADS and related models that uses modern Python libraries to achieve intuitive and user-friendly development and debugging. We describe how the code base allows rapid iteration and present results validating the implementation against its predecessors.

\section{The \texttt{lfads-torch} Library: An Overview}
Basic advantages of \texttt{lfads-torch} over previous implementations are eager execution, improved modularity, and less engineering boilerplate code. The advanced functionality spans three main categories: \textit{modules} (\S\ref{modules}), \textit{configuration} (\S\ref{configs}), and \textit{large-scale runs} (\S\ref{lgscale}).

\subsection{Modules}\label{modules}
\subsubsection{\texttt{augmentations} and the \texttt{AugmentationStack}}
Data augmentation is an effective tool for training LFADS models. In particular, augmentations that act on both the input data and the reconstruction cost gradients enable resistance to identity overfitting (coordinated dropout; \cite{keshtkaran2019enabling, keshtkaran2022large}) and the ability to infer firing rates with spatiotemporal superresolution (selective backpropagation through time \cite{zhu2021deep, zhu2022deep}). Other augmentations can reduce the impact of correlated noise \cite{wimalasena2022estimating}.

In \texttt{lfads-torch}, we provide a simple interface for applying data augmentations via the \texttt{AugmentationStack} class. The user creates the object by passing in a list of transformations and specifying the order in which they should be applied to the data batch, the loss tensor, or both. Separate \texttt{AugmentationStack}s are applied automatically by the \texttt{LFADS} object during training and inference, making it much easier to experiment with new augmentation strategies. Notably, the \texttt{augmentations} module provides implementations of \texttt{CoordinatedDropout} \cite{keshtkaran2019enabling, keshtkaran2022large}, \texttt{SelectiveBackpropThruTime} \cite{zhu2021deep}, and \texttt{TemporalShift} \cite{wimalasena2022estimating}.

\subsubsection{\texttt{priors}: Modular priors}
The LFADS model computes KL penalties between posteriors and priors for both initial condition and inferred input distributions, which are added to the reconstruction cost in the variational ELBO \cite{sussillo2016lfads, pandarinath2018inferring}. In the original implementation, priors were multivariate normal and autoregressive multivariate normal for the initial condition and inferred inputs, respectively. As priors impact the form of inferred inputs, users may want to experiment with alternate distributions that are more appropriate for certain brain areas and tasks.

With \texttt{lfads-torch}, users can implement custom priors by writing \texttt{make\_posterior} and \texttt{forward} functions and passing the modules to the \texttt{LFADS} constructor. In addition to the standard \texttt{MultivariateNormal} and \texttt{AutoregressiveMultivariateNormal} priors, we provide a \texttt{MultivariateStudentT} prior to encourage inference of heavy-tailed inputs \cite{schimel2021ilqr}.

\subsubsection{\texttt{recons}: Modular reconstruction}
The original LFADS implementation provided Poisson and Gaussian observation models. The former is often applied to binned spike counts and the latter has proven useful for electrocorticography \cite{flint2020representation}. Recent work has benefited from changing the output distribution for different data modalities \cite{wimalasena2022estimating, zhu2022deep}.

In \texttt{lfads-torch}, we include implementations of \texttt{Poisson}, \texttt{Gaussian}, \texttt{Gamma} \cite{wimalasena2022estimating}, and \texttt{ZeroInflatedGamma} \cite{zhu2022deep} distributions, widening the applicability of the implementation to EMG, calcium imaging, and more. We also provide an interface for users to implement new observation models by subclassing the abstract \texttt{Reconstruction} class.

Another noteworthy change is the separation of the data used as input (\texttt{encod\_data}) from the data being reconstructed (\texttt{recon\_data}). This enables applications like co-smoothing \cite{pei2021neural}, forward prediction \cite{pei2021neural}, and reconstruction across data modalities.

\subsubsection{Other PyTorch Lightning capabilities}
By building \texttt{lfads-torch} within the PyTorch Lightning ecosystem, we give users access to a large and growing array of complex functionality that would be technically challenging or tedious to implement manually. Examples include logging, checkpointing, profiling, mixed precision training, learning rate scheduling, pruning, accelerators, and more (see \href{https://pytorch-lightning.readthedocs.io/}{the docs}).

\subsection{Configuration}\label{configs}
In \texttt{lfads-torch}, a flexible configuration process gives users significant control over model and architecture configuration without needing to edit source code. Run configurations are built by composing a main YAML config file with selected \texttt{model}, \texttt{datamodule}, and \texttt{callbacks} configs. The target classes and arguments specified in these configs are directly used by Hydra to recursively instantiate the various objects used in the training pipeline. We provide several preprocessed datasets and example configurations for the Neural Latents Benchmark (NLB) \cite{pei2021neural} to help users understand the intended workflow.

\subsection{Large-scale runs}\label{lgscale}
While the \texttt{LFADS} class and other training objects can be instantiated directly like any other Python class, we provide flexible machinery for instantiating and training many models in parallel with different configurations for hyperparameter tuning. The \texttt{run\_model} function uses a config path, along with a set of overrides, to instantiate training objects and run the training loop. We provide \texttt{run\_single.py} for training single models in this manner, as well as \texttt{run\_multi.py} and \texttt{run\_pbt.py} for specifying hyperparameter spaces and performing large-scale searches and Population-based Training \cite{keshtkaran2022large} with Tune.

For users who want to perform large-scale hyperparameter tuning using AutoLFADS \cite{keshtkaran2022large} with minimal setup and infrastructure costs, we made \texttt{lfads-torch} available on NeuroCAAS \cite{abe2022neuroscience} for a convenient drag-and-drop modeling workflow.

\begin{wrapfigure}{r}{0.5\textwidth}
  \vspace{-50pt}
  \centering
  \includegraphics[width=0.48\textwidth]{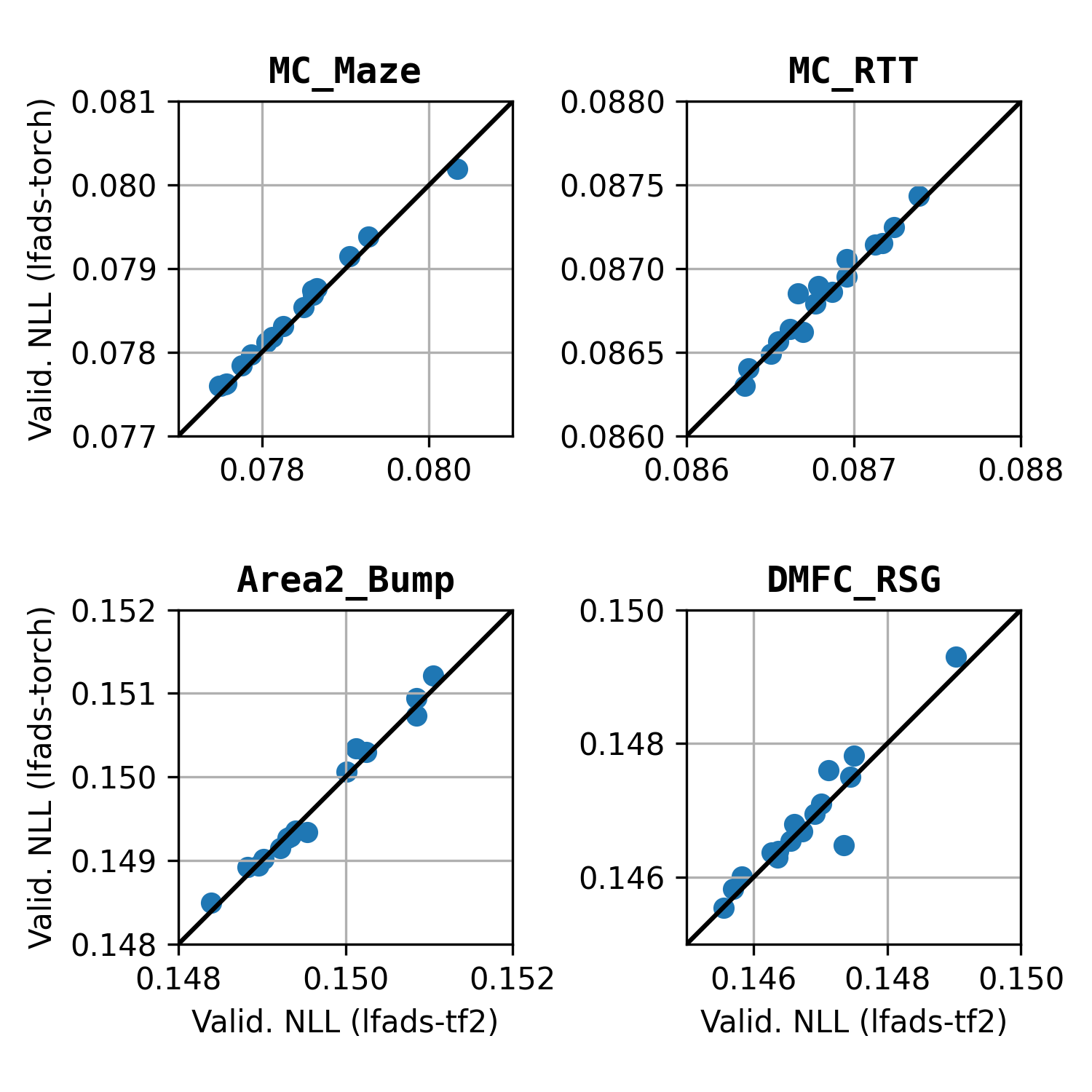}
  \caption{Validation loss for paired \texttt{lfads-tf2} and \texttt{lfads-torch} runs across four NLB datasets.}
  \vspace{-40pt}
  \label{fig:paired}
\end{wrapfigure}

\section{Validation}
We first verified that forward and backward passes of \texttt{lfads-torch} were identical to a previous implementation \cite{sedler2022} in the deterministic case. We then trained paired models with identical hyperparameters and initializations on several NLB datasets (5 ms; \texttt{MC\_Maze}, \texttt{MC\_RTT}, \texttt{Area2\_Bump}, \texttt{DMFC\_RSG}; Figure \ref{fig:paired}) and found that final losses were nearly identical across the hyperparameter space. Finally, we replicated performance of the NLB AutoLFADS baseline across all 5 ms datasets (see Table \ref{table:leaderboard}, Appendix Table \ref{table:leaderboard2} and \hyperlink{https://eval.ai/web/challenges/challenge-page/1256/leaderboard}{the EvalAI Leaderboard}).


\section{Summary}
\texttt{lfads-torch} is an implementation of LFADS that leverages modern deep learning frameworks, allowing easier application, robust hyperparameter tuning, and experimentation with new architectures and training methods. We hope that \texttt{lfads-torch} helps researchers push the boundaries of high-performance neural population modeling.

\begin{table}[h]
\begin{tabular}{|l|l|c|c|c|c|c|}
\hline
\multicolumn{1}{|c|}{Dataset} & \multicolumn{1}{c|}{Implementation} & co-bps          & vel R2          & psth R2         & fp-bps          & tp-corr          \\ \hline
\texttt{MC\_Maze}             & \texttt{lfads-tf2}                  & 0.3364          & \textbf{0.9097} & \textbf{0.6360} & 0.2349          &                  \\
             & \texttt{lfads-torch}                & \textbf{0.3497} & 0.9027          & 0.6170          & \textbf{0.2447} &                  \\ \hline
\texttt{MC\_RTT}              & \texttt{lfads-tf2}                  & 0.1868          & 0.6167          &                 & 0.1213          &                  \\
              & \texttt{lfads-torch}                & \textbf{0.1882} & \textbf{0.6176} &                 & \textbf{0.1240} &                  \\ \hline
\texttt{Area2\_Bump}          & \texttt{lfads-tf2}                  & \textbf{0.2569} & 0.8492          & \textbf{0.6318} & \textbf{0.1505} &                  \\
          & \texttt{lfads-torch}                & 0.2535          & \textbf{0.8516} & 0.6104          & 0.1455          &                  \\ \hline
\texttt{DMFC\_RSG}            & \texttt{lfads-tf2}                  & \textbf{0.1829} &                 & \textbf{0.6359} & 0.1844          & \textbf{-0.8248} \\
            & \texttt{lfads-torch}                & 0.1820          &                 & 0.5873          & \textbf{0.1960} & -0.7627          \\ \hline
\end{tabular}
\caption{Test performance of \texttt{lfads-tf2} and \texttt{lfads-torch} for four NLB datasets (5 ms).}\label{table:leaderboard}
\vspace{-20pt}

\end{table}

\acks{This work was supported by NSF Graduate Research Fellowship DGE-2039655 (ARS), NIH-NINDS/OD DP2NS127291, NIH BRAIN/NIDA RF1 DA055667, and the Alfred P. Sloan Foundation (CP). We declare no competing interests.}

\vskip 0.2in
\bibliography{references}

\newpage
\section{Appendix}

\begin{table}[h]
\begin{tabular}{|l|l|c|c|c|c|c|}
\hline
\multicolumn{1}{|c|}{Dataset} & \multicolumn{1}{c|}{Implementation} & co-bps          & vel R2          & psth R2         & fp-bps          & tp-corr          \\ \hline
\texttt{MC\_Maze}             & \texttt{lfads-tf2}                  & 0.3364          & \textbf{0.9097} & \textbf{0.6360} & 0.2349          &                  \\
             & \texttt{lfads-torch}                & \textbf{0.3497} & 0.9027          & 0.6170          & \textbf{0.2447} &                  \\ \hline
\texttt{MC\_RTT}              & \texttt{lfads-tf2}                  & 0.1868          & 0.6167          &                 & 0.1213          &                  \\
              & \texttt{lfads-torch}                & \textbf{0.1882} & \textbf{0.6176} &                 & \textbf{0.1240} &                  \\ \hline
\texttt{Area2\_Bump}          & \texttt{lfads-tf2}                  & \textbf{0.2569} & 0.8492          & \textbf{0.6318} & \textbf{0.1505} &                  \\
          & \texttt{lfads-torch}                & 0.2535          & \textbf{0.8516} & 0.6104          & 0.1455          &                  \\ \hline
\texttt{DMFC\_RSG}            & \texttt{lfads-tf2}                  & \textbf{0.1829} &                 & \textbf{0.6359} & 0.1844          & \textbf{-0.8248} \\
            & \texttt{lfads-torch}                & 0.1820          &                 & 0.5873          & \textbf{0.1960} & -0.7627          \\ \hline
\texttt{MC\_Maze\_Small}      & \texttt{lfads-tf2}                  & 0.2854          & \textbf{0.7982} & 0.3342          & 0.1244          &                  \\
      & \texttt{lfads-torch}                & \textbf{0.2988} & 0.7834          & \textbf{0.4229} & \textbf{0.1346} &                  \\ \hline
\texttt{MC\_Maze\_Medium}     & \texttt{lfads-tf2}                  & \textbf{0.3036} & 0.8680          & \textbf{0.5969} & \textbf{0.1636} &                  \\
     & \texttt{lfads-torch}                & 0.2979          & \textbf{0.8926} & 0.5935          & 0.1609          &                  \\ \hline
\texttt{MC\_Maze\_Large}      & \texttt{lfads-tf2}                  & \textbf{0.3740} & 0.9178          & \textbf{0.7261} & \textbf{0.2039} &                  \\
      & \texttt{lfads-torch}                & 0.3710          & 0.9178          & 0.7190          & 0.2000          &                  \\ \hline
\end{tabular}
\caption{Test performance of \texttt{lfads-tf2} and \texttt{lfads-torch} across all 5 ms NLB datasets.}
\label{table:leaderboard2}
\end{table}

\end{document}